\documentclass[10pt, a4paper]{article}

\usepackage[utf8]{inputenc}
\usepackage[T1]{fontenc}
\usepackage{graphicx}
\usepackage{amsmath}
\usepackage{amsfonts}
\usepackage{amssymb}
\usepackage{geometry}
\usepackage{hyperref}
\usepackage[english]{babel}

\usepackage{booktabs}       
\usepackage[svgnames]{xcolor} 
\usepackage{caption}        
\usepackage{subcaption}     
\usepackage{float}          
\usepackage{listings}       
\usepackage{authblk}        
\usepackage{url}

\hypersetup{
    colorlinks=true,
    linkcolor=blue,
    filecolor=magenta,      
    urlcolor=cyan,
    pdftitle={Hierarchical Deep Fusion Framework for Multi-dimensional Facial Forgery Detection},
    pdfpagemode=FullScreen,
}

\definecolor{codegreen}{rgb}{0,0.6,0}
\definecolor{codegray}{rgb}{0.5,0.5,0.5}
\definecolor{codepurple}{rgb}{0.58,0,0.82}
\definecolor{backcolour}{rgb}{0.95,0.95,0.92}

\lstdefinestyle{mystyle}{
    backgroundcolor=\color{backcolour},   
    commentstyle=\color{codegreen},
    keywordstyle=\color{magenta},
    numberstyle=\tiny\color{codegray},
    stringstyle=\color{codepurple},
    basicstyle=\ttfamily\footnotesize,
    breakatwhitespace=false,         
    breaklines=true,                 
    captionpos=b,                    
    keepspaces=true,                 
    numbers=left,                    
    numbersep=5pt,                  
    showspaces=false,                
    showstringspaces=false,
    showtabs=false,                  
    tabsize=2
}
\title{\textbf{Hierarchical Deep Fusion Framework for Multi-dimensional Facial Forgery Detection - The 2024
Global Deepfake Image Detection Challenge}}

\author{Kohou Wang}
\author{Huan Hu}
\author{Xiang Liu}
\author{Zezhou Chen}
\author{Ping Chen}
\author{Zhaoxiang Liu}
\author{Shiguo Lian}

\affil{AI Innovation Center, China Unicom}

\date{October 23, 2024} 

\begin{document}

\maketitle

\begin{abstract}
The proliferation of sophisticated deepfake technology poses significant challenges to digital security and authenticity. Detecting these forgeries, especially across a wide spectrum of manipulation techniques, requires robust and generalized models. This paper introduces the Hierarchical Deep Fusion Framework (HDFF), an ensemble-based deep learning architecture designed for high-performance facial forgery detection. Our framework integrates four diverse pre-trained sub-models, Swin-MLP, CoAtNet, EfficientNetV2, and DaViT, which are meticulously fine-tuned through a multi-stage process on the MultiFFDI dataset. By concatenating the feature representations from these specialized models and training a final classifier layer, HDFF effectively leverages their collective strengths. This approach achieved a final score of 0.96852 on the competition's private leaderboard, securing the 20th position out of 184 teams, demonstrating the efficacy of hierarchical fusion for complex image classification tasks.
\end{abstract}

\section{Introduction}
In recent years, advancements in generative artificial intelligence have led to the creation of highly realistic deepfakes, which present unprecedented threats to information integrity and personal security. The ability to manipulate facial images and videos through techniques like face swapping, attribute editing, and full synthesis necessitates the development of advanced detection systems. The Multi-dimensional Facial Forgery Detection challenge aims to address this issue by providing a comprehensive and diverse dataset, MultiFFDI, which includes a wide variety of forgery types generated by over 50 different methods.

Traditional detection methods often struggle to generalize across unseen forgery techniques. To overcome this limitation, ensemble learning has emerged as a powerful strategy, combining the predictions of multiple models to improve overall robustness and accuracy. However, simply averaging outputs may not be optimal. A more sophisticated approach is to fuse features from diverse models, allowing a final classifier to learn the complex interplay between different feature representations.

In this work, we propose the Hierarchical Deep Fusion Framework (HDFF), an ensemble architecture that systematically combines the strengths of four distinct pre-trained models. Our contributions are threefold:
\begin{itemize}
    \item We design a novel fusion architecture that integrates diverse backbones (Swin-MLP, CoAtNet, EfficientNetV2, and DaViT) to capture a wide range of forgery artifacts.
    \item We introduce a multi-stage training strategy, involving selective and then comprehensive fine-tuning of sub-models, followed by training a dedicated fusion layer, which stabilizes learning and maximizes performance.
    \item We demonstrate the effectiveness of our framework, achieving a high rank in a competitive challenge and providing insights into practical model development, such as the importance of full-network fine-tuning and advanced learning rate schedulers.
\end{itemize}

\section{Proposed Method}
\begin{figure}[H]
    \centering
    \includegraphics[width=\linewidth]{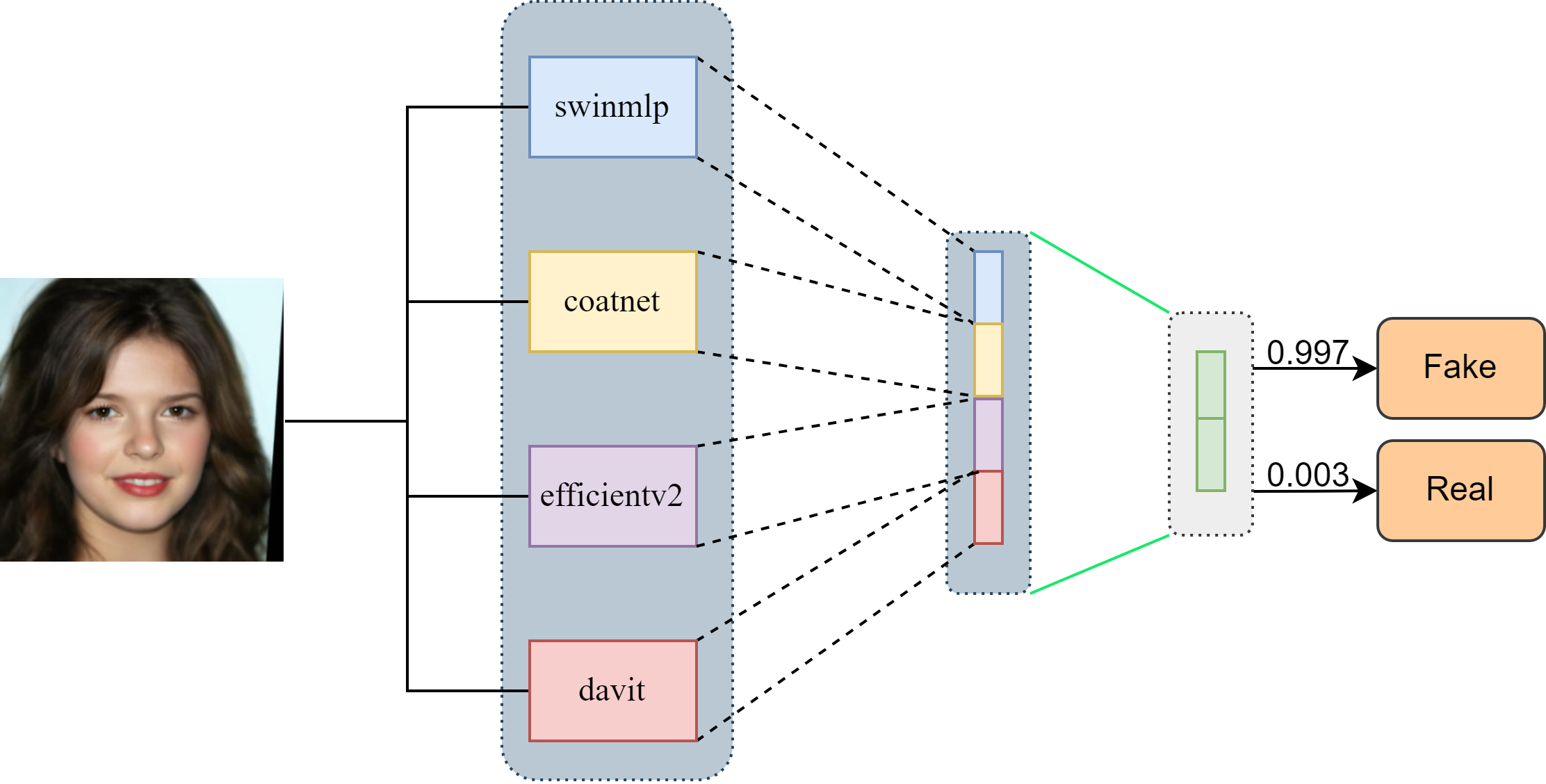} 
    \caption{The architecture of our Hierarchical Deep Fusion Framework (HDFF). Four pre-trained sub-models are individually fine-tuned. Their final feature layers are then concatenated and fed into a new fully connected layer, which serves as the ultimate classifier.}
    \label{fig:hdff_architecture}
\end{figure}

Our solution, the Hierarchical Deep Fusion Framework (HDFF), is built upon the principle of model ensembling. Instead of relying on a single model, we leverage a collection of specialized sub-models, whose outputs are intelligently combined to make a final, more accurate prediction. The overall workflow of our framework is depicted in Figure~\ref{fig:hdff_architecture}.

\subsection{Model Architecture and Fusion Strategy}
The core of HDFF consists of four powerful and diverse sub-models, each pre-trained on the ImageNet-1K dataset: \textbf{Swin-MLP}~\cite{zheng2022swin}, \textbf{CoAtNet}~\cite{dai2021coatnet}, \textbf{EfficientNetV2}~\cite{tan2021efficientnetv2}, and \textbf{DaViT}~\cite{ding2022davit}. The diversity in their architectures—spanning from pure transformers and MLPs to hybrid convolution-attention models—allows them to capture different types of forgery artifacts.

The fusion process is as follows:
\begin{enumerate}
    \item Each fine-tuned sub-model acts as a feature extractor.
    \item We take the output of the final layer (before the original classifier) from each sub-model.
    \item These feature vectors are concatenated into a single, larger vector.
    \item A new, fully connected (FC) layer is appended to this concatenated vector. This FC layer functions as the "grand model's" classifier, learning to weigh the features from each sub-model to produce the final classification.
\end{enumerate}

\begin{figure}[H]
    \centering
    \includegraphics[width=\linewidth]{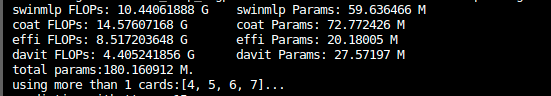} 
    \caption{The parameter distribution of the sub-models within our HDFF. The total size is 180.16M, satisfying the competition's 200MB constraint.}
    \label{fig:total_params}
\end{figure}

A critical constraint of the challenge was a model size limit of 200MB. We meticulously managed the parameters of our ensemble to stay within this budget. As shown in Figure~\ref{fig:total_params}, the total parameter count of our final model is 180.16M, comfortably below the limit.

\subsection{Data Preprocessing and Augmentation}
To enhance model robustness and generalization, we applied a standardized preprocessing and augmentation pipeline to all input images. The primary augmentation strategy employed was \textbf{AutoAugment}~\cite{cubuk2018autoaugment}, which applies a learned sequence of augmentation policies to the data. Our transformation pipeline is outlined in the PyTorch-style code in Listing~\ref{lst:data_transform}.

\begin{lstlisting}[language=Python, caption={Data preprocessing and augmentation pipeline.}, label=lst:data_transform]
from torchvision import transforms

input_size = 224 # Example input size
transform = transforms.Compose([
    transforms.Resize((input_size, input_size)),
    AutoAugment(), # Applies learned policies
    transforms.ToTensor(),
    transforms.Normalize(mean=[0.485, 0.456, 0.406],
                         std=[0.229, 0.224, 0.225]),
])
\end{lstlisting}

The AutoAugment policies include a variety of transformations like `Invert`, `Rotate`, `Sharpness`, `ShearY`, `TranslateX`, `Color`, and `Brightness`, applied with stochastic probabilities and magnitudes. This forces the model to learn features that are invariant to these transformations, preventing overfitting and improving performance on unseen data.

\subsection{Multi-Stage Training Strategy}
A key component of our method is the carefully designed multi-stage training strategy, which ensures that both the sub-models and the final fusion layer are optimally trained.

\begin{enumerate}
    \item \textbf{Sub-Model Initialization}: Each of the four sub-models is initialized with its weights pre-trained on ImageNet-1K. This provides a strong foundation for generic feature extraction.

    \item \textbf{Selective Fine-Tuning}: Initially, we freeze all layers of each sub-model except for the final fully connected (classifier) layer. This layer is then trained on the MultiFFDI dataset. This step allows the model to quickly adapt its classification head to the new domain without disrupting the learned feature hierarchies.

    \item \textbf{Comprehensive Fine-Tuning}: After the classifier head has adapted, we unfreeze all layers of each sub-model and continue training on the MultiFFDI dataset with a smaller learning rate. This end-to-end fine-tuning allows the entire network to specialize in detecting subtle forgery artifacts specific to the dataset.

    \item \textbf{Grand Model Training}: Once all sub-models are individually optimized, they are integrated into the HDFF architecture. We freeze the weights of all four sub-models and train \textit{only} the newly added final FC fusion layer. This final step is efficient and effective, as it solely focuses on learning how to best combine the already powerful features extracted by the sub-models.
\end{enumerate}

\section{Experiments and Results}

\subsection{Implementation Details}
Our framework was implemented using Python and the PyTorch deep learning library. All training and inference were conducted on a high-performance computing system equipped with four NVIDIA A100 (80GB) GPUs. We utilized the AdamW optimizer with a \textbf{CosineAnnealingLR}~\cite{loshchilov2016sgdr} scheduler, which cyclically adjusts the learning rate. This scheduler proved to be more effective than a standard StepLR schedule. Each sub-model required approximately 2-3 days for comprehensive fine-tuning, with the final grand model training taking an additional 2-3 days to converge to optimal performance.

\subsection{Performance Evaluation}
Our proposed Hierarchical Deep Fusion Framework demonstrated strong performance in the challenge. The final submission achieved a score of \textbf{0.9685225726} on the private leaderboard, securing the \textbf{20th rank out of 184 participating teams}. This result validates the effectiveness of our multi-stage training and hierarchical fusion approach for the complex task of multi-dimensional facial forgery detection.

\subsection{Ablation Studies and Key Observations}
During the development process, we made several critical observations that guided our final design.

\paragraph{Efficacy of Cosine Annealing Learning Rate.}
We compared the performance of a standard StepLR scheduler with the CosineAnnealingLR scheduler. The cosine annealing schedule, which smoothly varies the learning rate in a sinusoidal fashion, provided a significant performance boost. On the phase 1 test set, this change alone improved the accuracy of a single model from 0.9949 to 0.9967, highlighting the importance of advanced optimization techniques.

\paragraph{The Necessity of Full-Network Fine-Tuning.}
Our initial experiments followed a common transfer learning practice of only training the final fully connected layer. However, this severely limited the model's potential. For instance, our SwinMLP model's accuracy was capped at approximately 0.75 when only its final layer was trained. We initially perceived this as a failure of the model choice. However, upon deciding to unfreeze all parameters and perform comprehensive fine-tuning (Stage 3 of our strategy), the model's performance dramatically improved, eventually reaching competitive accuracy levels. This experience underscores a crucial lesson: for complex and domain-specific tasks like forgery detection, adapting the entire feature extraction backbone is often necessary to achieve optimal results.

\section{Conclusion}
In this paper, we presented the Hierarchical Deep Fusion Framework (HDFF), a robust ensemble method for detecting multi-dimensional facial forgeries. By combining four diverse pre-trained models and employing a meticulous multi-stage fine-tuning strategy, our framework effectively captures a wide array of forgery artifacts. The final fusion mechanism, which trains a dedicated classifier on the concatenated features of the sub-models, proved to be a highly effective approach.

Our final ranking of 20th in a competitive field affirms the strength of our methodology. Furthermore, our experiments provided valuable insights, such as the significant impact of the learning rate scheduler and the critical need for full-network fine-tuning in transfer learning for specialized domains.

Looking ahead, we believe there is still room for improvement. A promising direction for future work is the exploration of multi-modal features. For example, predicting depth maps from facial images and analyzing them for inconsistencies could provide a powerful, orthogonal signal for distinguishing real faces from synthetic ones. While time constraints prevented us from exploring this avenue during the competition, it remains a compelling area for future research in the fight against deepfakes.



\begin{thebibliography}{9}

\bibitem{cubuk2018autoaugment}
Cubuk, E. D., Zoph, B., Mane, D., Vasudevan, V., \& Le, Q. V. (2018).
\textit{Autoaugment: Learning augmentation policies from data}.
arXiv preprint arXiv:1805.09501.

\bibitem{zheng2022swin}
Zheng, H., Wang, G., \& Li, X. (2022).
\textit{Swin-MLP: A strawberry appearance quality identification method by Swin Transformer and multi-layer perceptron}.
Journal of Food Measurement and Characterization, 16(4), 2789-2800.

\bibitem{dai2021coatnet}
Dai, Z., Liu, H., Le, Q. V., \& Tan, M. (2021).
\textit{Coatnet: Marrying convolution and attention for all data sizes}.
Advances in neural information processing systems, 34, 3965-3977.

\bibitem{tan2021efficientnetv2}
Tan, M., \& Le, Q. (2021, July).
\textit{Efficientnetv2: Smaller models and faster training}.
In International conference on machine learning (pp. 10096-10106). PMLR.

\bibitem{ding2022davit}
Ding, M., Xiao, B., Codella, N., Luo, P., Wang, J., \& Yuan, L. (2022, October).
\textit{Davit: Dual attention vision transformers}.
In European conference on computer vision (pp. 74-92). Cham: Springer Nature Switzerland.

\bibitem{loshchilov2016sgdr}
Loshchilov, I., \& Hutter, F. (2016).
\textit{Sgdr: Stochastic gradient descent with warm restarts}.
arXiv preprint arXiv:1608.03983.

\end{thebibliography}
\end{document}